\documentclass[a4paper]{article}
\pdfoutput=1

\usepackage[english]{babel}
\usepackage[utf8]{inputenc}
\usepackage[T1]{fontenc}

\usepackage{amsfonts,amssymb,amsmath,amsthm}
\usepackage{graphicx}
\usepackage[footnotesize]{caption}
\usepackage[position=b,justification=Centering,singlelinecheck=false]{subfig}
\usepackage[usenames,dvipsnames]{xcolor}
\usepackage{pgfplots}
\pgfplotsset{compat=newest}
\usepackage{pgfplotstable}
\usepackage{tikz}	
\usepackage{empheq}
\usepackage[preprint]{spconf-arxiv}
\graphicspath{{pics/}}		

\toappear{To appear in {\it Proc.\ iTWIST2016,
                 August 24 - 26, 2016, Aalborg, Denmark}}

\newcommand{\bs}{\boldsymbol}
\newcommand{\mbf}{\mathbf}
\newcommand{\bb}{\mathbb}
\newcommand{\cl}{\mathcal}
\newcommand{\ie}{\emph{i.e.}, }
\newcommand{\eg}{\emph{e.g.}, }
\DeclareMathOperator*{\argmin}{\arg\min}

\newlength{\sizeFig}
\newlength{\hspaceFig}
\newlength{\vspaceSec}
\newlength{\vspaceSecD}
\newlength{\vspaceEq}
\title{Blind Deconvolution of PET Images using Anatomical Priors}
\name{St\'ephanie Gu\'erit$^{1,*}$\ , Adriana Gonz\'alez$^{1,*}$, Anne Bol$^2$, John A. Lee$^{1,2,\dagger}$ \ and Laurent Jacques$^{1,\dagger}$%
	\thanks{$^*$ These authors contributed equally.}%
	\thanks{$^\dagger$ Research Associates with the Belgian F.R.S.-FNRS.}%
}
\address{
	\normalsize $^1$ ICTEAM/ELEN, ISPGroup, $^2$ MIRO/IREC, Universit\'e catholique de Louvain, Belgium\\
}
\begin{document}

\maketitle
\begin{abstract}
Images from positron emission tomography (PET) provide metabolic information about the human body. They present, however, a spatial resolution that is limited by physical and instrumental factors often modeled by a blurring function. Since this function is typically unknown, blind deconvolution (BD) techniques are needed in order to produce a useful restored PET image.
In this work, we propose a general BD technique that restores a low resolution blurry image using information from data acquired with a high resolution modality (\eg CT-based delineation of regions with uniform activity in PET images). The proposed BD method is validated on synthetic and actual phantoms.
\end{abstract}

\vspace{\vspaceSec}
\section{Introduction}
\label{sec:introduction}

Positron emission tomography (PET) is a powerful functional imaging technique. Often combined with anatomical computed tomography (CT) images, it provides physicians with relevant information for patient management~\cite{Bailey2005}. For instance, a radioactive analogue of glucose, the $^{18}$F-fluorodeoxyglucose ($^{18}$FDG), is injected in the patient's body and accumulates mainly in tissues of abnormally high metabolic activity. Emitted positrons are subsequently detected by the PET imaging system via anti-collinear annihilation photons.

\subsection{Motivation}
PET data suffer from two drawbacks restricting their direct use: \emph{(i)} low spatial resolution and \emph{(ii)} high level of noise. Both physical and instrumental factors limit the spatial resolution~\cite{Bailey2005}. They are commonly represented as a blurring function that is estimated by imaging a radioactive point source. The resulting blur is usually approximated by a Gaussian point spread function (PSF)~\cite{Bailey2005}. Noise, which is Poissonian in the raw data, is due to low photon detection efficiency of the detectors and to the limited injected tracer dose. 

Improving the quality of PET images is a key element for better medical interpretation. Post-reconstruction restoration techniques are more adapted to clinical use since only reconstructed images are generally available. Nowadays, existing approaches are dedicated to denoising, deblurring, or combining both steps~\cite{Lee2008, ARCS2006, Sroubek2009, Guerit2015}. Most of the approaches that include deblurring use an empirically estimated PSF.

Since the design of PET scanners can vary greatly between manufacturers and models, their imaging properties (\eg their PSF) are different. Restoration of images acquired in the context of a multicentric study (with centralization of the reconstructed data) is then challenging: the raw data are not always accessible and the PSF cannot be directly measured. This context motivates the development of \emph{blind} deconvolution (BD) techniques to jointly estimate the PSF and the PET image.
\vspace{\vspaceSec}
\subsection{Contributions}
\begin{figure}
	\centering
		\subfloat{\includegraphics[height=1.5\sizeFig]{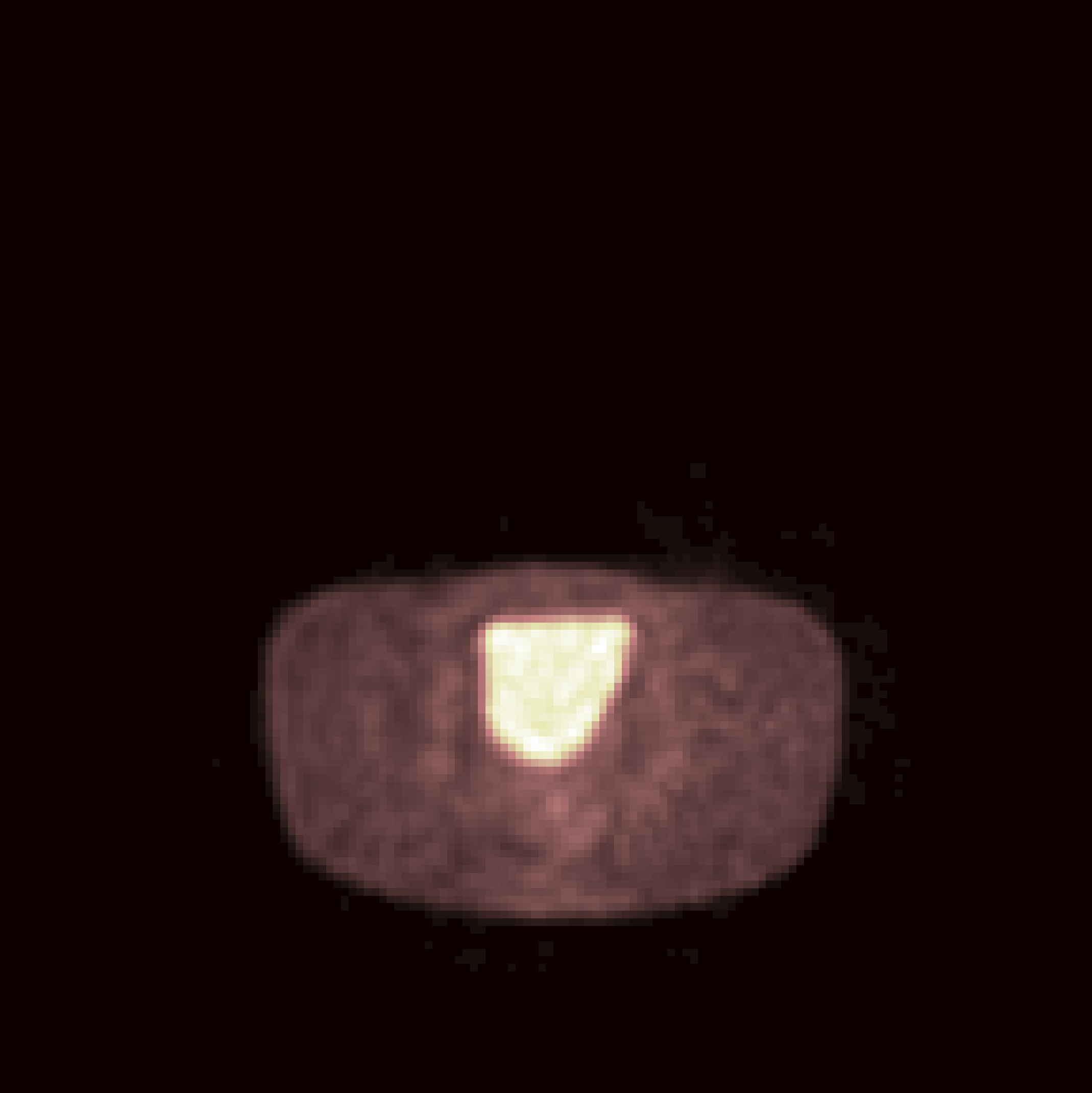}} \hspace{\hspaceFig}
		\subfloat{\includegraphics[height=1.5\sizeFig]{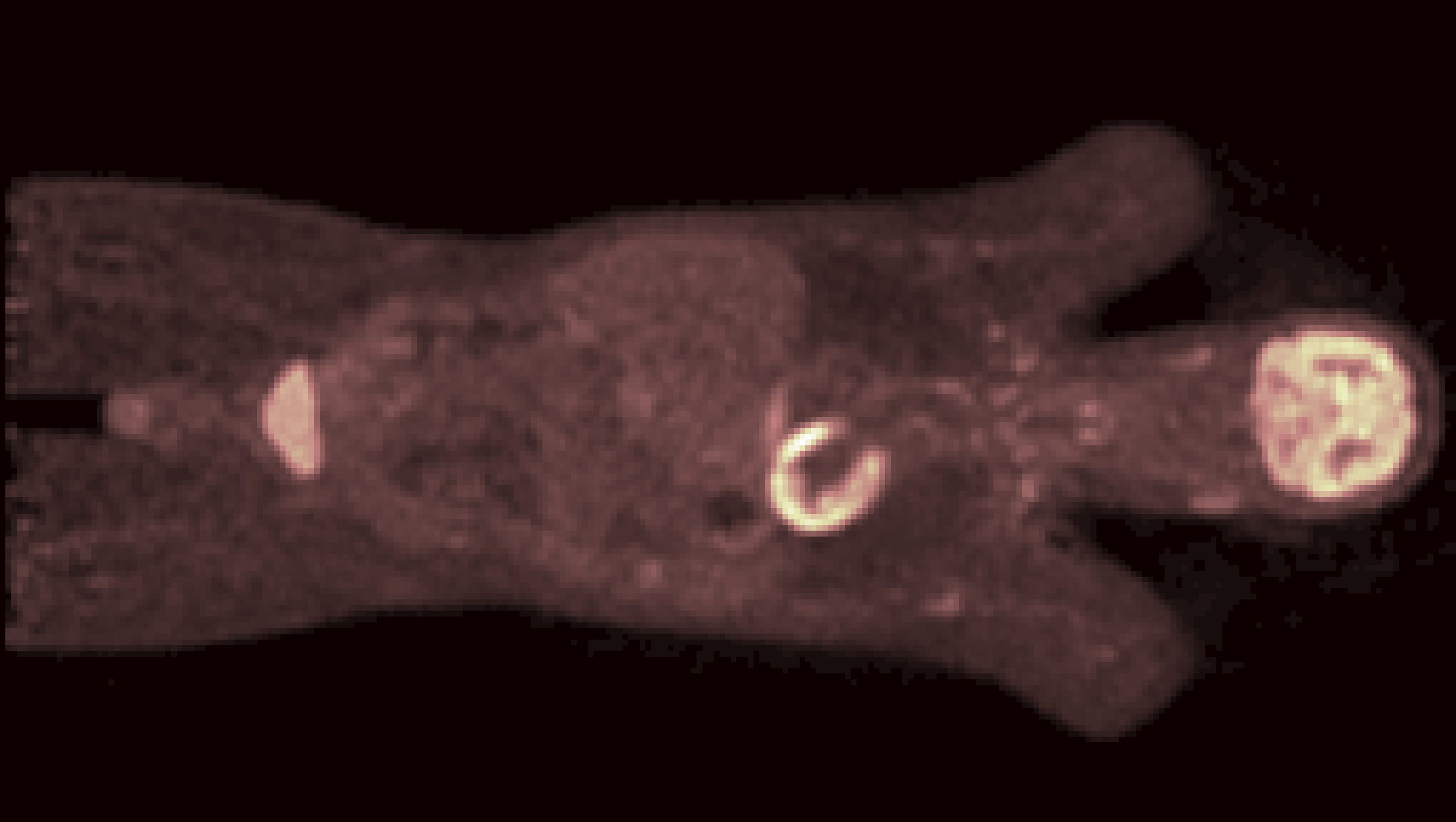}} \hspace{\hspaceFig}
	\caption{Transaxial (left) and coronal (right) PET images of a patient. Activity in the bladder is high and uniform. Source: Cliniques universitaires Saint-Luc.}
	\label{fig:bladder}
\end{figure}

Anatomical information can be useful to regularize such an ill-posed inverse problem~\cite{Sroubek2009}. Due to its perfect dilution in the kidneys, $^{18}$FDG accumulates uniformly in the patient bladder with a high concentration (see Fig.~\ref{fig:bladder})~\cite{Bailey2005}.  An accurate delineation of this organ is obtained from high resolution CT images. This provides a strong prior information that can be used to regularize the inverse problem: inside this region, the intensity of the restored image must be constant (\ie the gradient is zero). In this work, we propose a method using this prior to regularize the BD inverse problem. A similar prior is used in~\cite{GDJ2016} for BD of astronomical images with a celestial transit. Any PSF is directly inferred from the observations without assuming a parametric model.
We only assume that the PSF \emph{(i)}~preserves the photon counts, \emph{(ii)} is non-negative and \emph{(iii)} is spatially invariant over the whole field-of-view (FOV). 

\vspace{\vspaceSec}
\section{Problem Statement}
\label{sec:prob}

Let $\mbf x \in \bb R^N$ be the original PET image defined in a $n \times n$ grid of $N = n^2$ pixels. Since the PSF is assumed to be uniform, the observed PET image $\mbf y \in \bb R^N$ can be modeled as a linear convolution of $\mbf x$ with a function $\mbf h \in \bb R^N$, the discrete equivalent of the PSF. 
If we assume that the acquisition time is long enough to have a high photon counts, the Poisson noise intrinsic to the PET imaging process can be approximated as an additive, white and Gaussian noise (AWGN) $\bs \eta \in \bb R^{N}$, with $\eta_i \sim_{\rm i.i.d.} \cl N(0,\sigma_\text{n}^2)$. The acquisition process is thus modeled as follows:
\begin{equation}
\label{eq:discrete_model_vector}
	\mbf y = \mbf h \otimes \mbf x + \bs \eta,
\end{equation}
where $\| \bs \eta \|_2^2 < \varepsilon^2 \coloneqq \sigma_\text{n}^2 (N + c\sqrt{N})$ with high probability for $c = \cl O (1)$~\cite{Hoeffding1963}. Noise variance $\sigma_\text{n}^2$ can be estimated using a robust median estimator~\cite{DJ1994}.
\vspace{\vspaceSec}
\section{Blind Deconvolution}
\label{sec:BID}

In this work, we aim at reconstructing both $\mbf x$ and $\mbf h$ from the noisy observations $\mbf y$. Since the data are assumed to be corrupted by AWGN, we adopt a data fidelity term based on the $\ell_2$-norm of the residual vector, \ie $ \| \mbf y - \mbf h \otimes \mbf x \|_2$. 

If we formulate the BD as a least-squares problem that minimizes the energy of the residual vector, we have an ill-posed problem since an infinite number of solutions can produce the observations $\mbf y$. The problem needs to be regularized by enforcing prior knowledge about the PSF and the image, avoiding in this way the trivial solution.

We use an anatomical prior where the image is assumed to be constant inside set $\Omega$ containing the pixels that represent the patient's bladder. This means that the image gradient belongs to the set of pixels with zero intensity in $\Omega$ as \mbox{$\mathcal{G} = \{\mbf u \in \mathbb{R}^{2N} | u_i = 0, \forall i \in \Omega\}$}.

As commonly done in the literature~\cite{Sroubek2009,AABM2012,CP2011}, we assume that images analyzed in this work are composed by slowly varying areas separated by sharp boundaries corresponding to tissue interfaces. Therefore, the inverse problem can be further regularized by promoting small Total-Variation norm (denoted $\| \mbf u \|_\text{TV}$), which corresponds to the $\ell_1$-norm of the magnitude of the gradient of $\mbf u \in \mathbb{R}^N$~\cite{CP2011}. 

The physics of PET imaging suggests three additional constraints to model~\eqref{eq:discrete_model_vector} : \emph{(i)} image \textit{non-negativity}, \ie $\mbf x \in \mathcal{P}=\{\mbf u \in \mathbb{R}^N | \mbf u \succeq \boldsymbol{0}\}$, since the original image $\mbf x$ represents non-negative metabolic activity; \emph{(ii)} \textit{photometry invariance}, \ie $\sum_{i=1}^{N} x_i \approx  \sum_{i=1}^{N} y_i$; \emph{(iii)} PSF \textit{non-negativity}, since the PSF is an observation of a point. From \emph{(ii)} and \emph{(iii)} we know that the PSF must belong to the probability simplex~\cite{PB2014}, defined as ${\cl PS} = \{ \mbf h \in \bb R^N | \mbf h \succeq \bs 0, \| \mbf h \|_1 = 1 \}$.

Gathering all these aspects, we propose the following regularized BD formulation:
\begin{equation}
\label{eq:BID}
	\{\widetilde{\mbf x}_\text{B}, \widetilde{\mbf h}_\text{B}\} = \argmin_{\overline{\mbf x}, \overline{\mbf h}} \ \{ L(\overline{\mbf x}, \overline{\mbf h}) \coloneqq \rho \| \overline{\mbf x} \|_\text{TV} + \textstyle\frac{1}{2} \| \mbf y - \overline{\mbf h} \otimes \overline{\mbf x} \|_2^2 + \imath_{\cl P} (\overline{\mbf x}) + \imath_{\cl G} (\bs \nabla \overline{\mbf x}) + \imath_{\cl {PS}} (\overline{\mbf h})\},
\end{equation}
with $\overline{\mbf x}, \ \overline{\mbf h} \in \bb R^N$, $L(\overline{\mbf x}, \overline{\mbf h})$ the objective function and \mbox{$\bs \nabla : \bb R^N \to \bb R^{2N}$} the gradient operator~\cite{Chambolle1994}. The function $\imath_{\cl C} (\mbf u)$ denotes the convex indicator function on the set $\cl C \in \{\cl P, \cl G, \cl PS \}$, which is equal to $0$ if $\mbf u \in \cl C$ and $+\infty$ otherwise. 
Regularization parameter $\rho$, unknown \emph{a priori}, controls the trade-off between sparsity of the image gradient magnitude and fidelity to the observations. In this work, we estimate $\rho$ iteratively based on~\cite{GDJ2016}, with an initial value given by $\sigma \sqrt{2 \log N}$~\cite{DJ1994}. Non-convex problem~\eqref{eq:BID} is solved by means of the proximal alternating minimization proposed in~\cite{ABRS2010}. 

Once the PSF has been estimated using the BD scheme~\eqref{eq:BID}, image $\widetilde{\mbf x}_\text{NBD}$ can be estimated using a non-blind deconvolution (NBD) scheme that does not take into account the anatomical prior. The NBD formulation is solved for an observation $\mbf y$ that has not been used in~\eqref{eq:BID} for the PSF estimation.


\vspace{\vspaceSec}
\section{Results and discussion}
\label{sec:exp}

\subsection{Method} 
\begin{figure}
	\centering
		\subfloat[\label{fig:results_synthetic_phantom_a}Original image $\mbf x$]{\includegraphics[height=1.1\sizeFig]{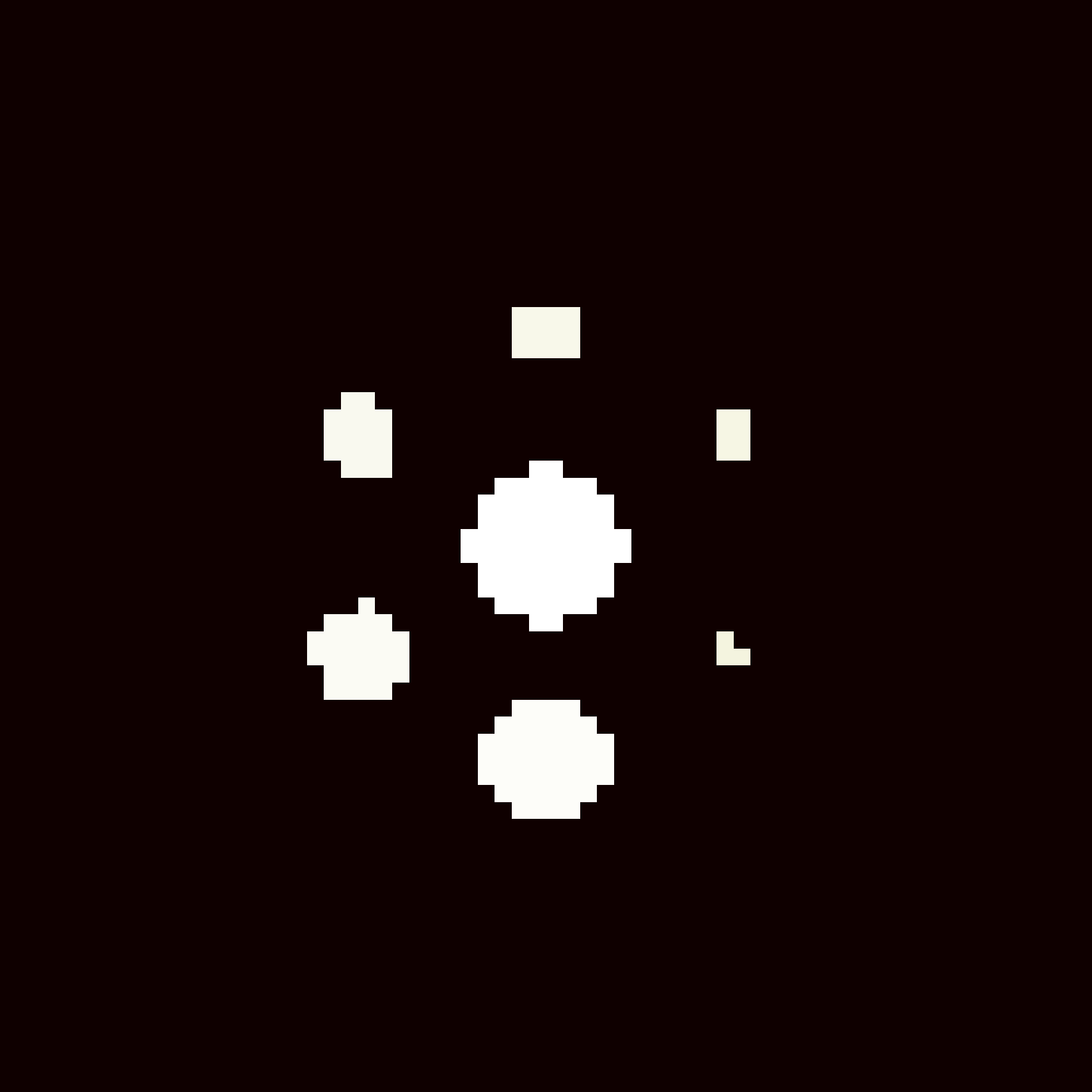}\begin{tikzpicture}[overlay]\draw [outer xsep=-2pt] (0,0) node [outer xsep=-2pt] {};\draw [cyan,very thick,dashed] (-0.5,1.15) -- (-1.85,1.15);	\end{tikzpicture}} \hspace{\hspaceFig}
		\subfloat[\label{fig:results_synthetic_phantom_b}Ground truth $\mbf h$]{\includegraphics[height=1.1\sizeFig]{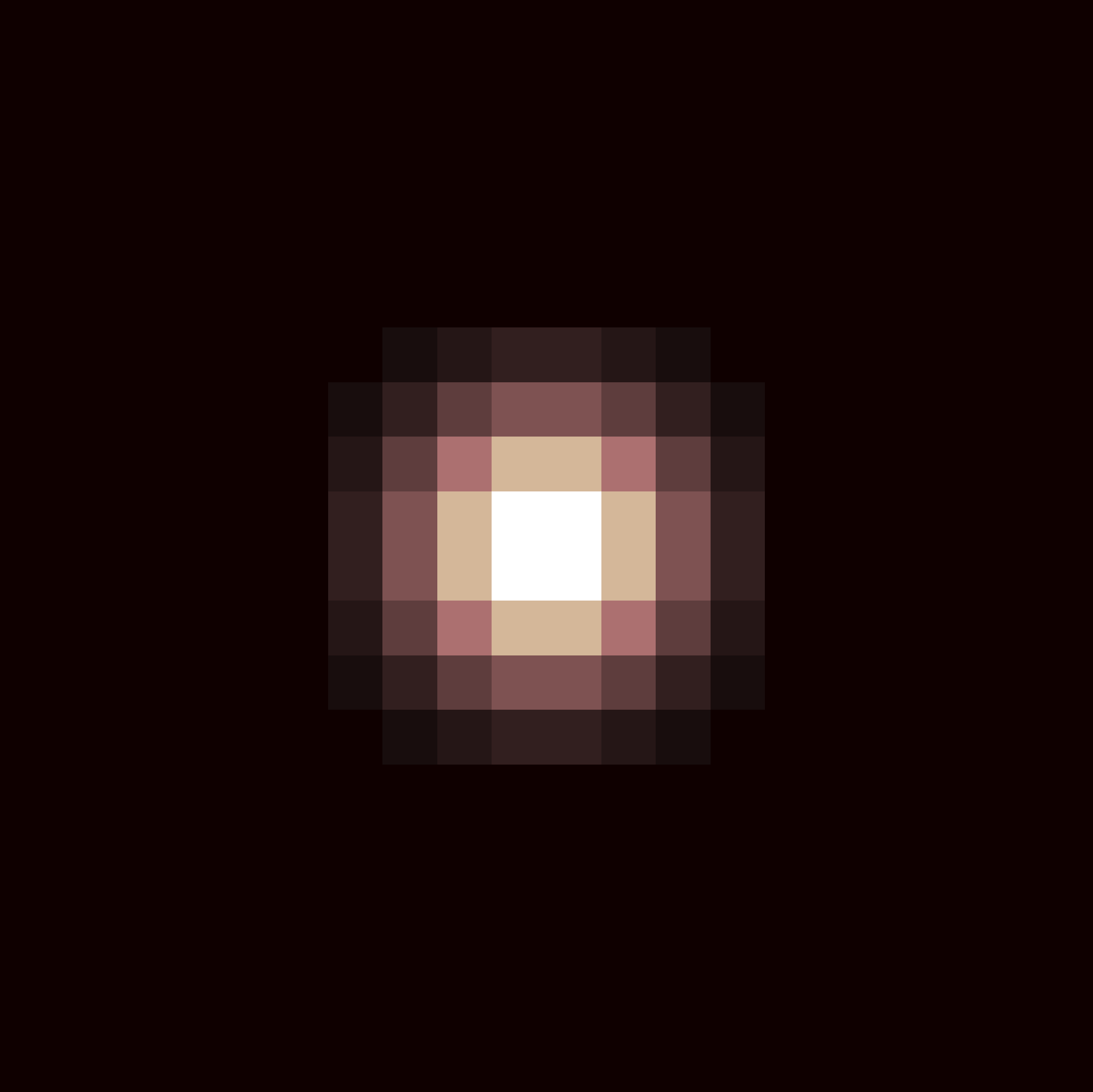}} \hspace{\hspaceFig}
		\subfloat[\label{fig:results_synthetic_phantom_c}Observations $\mbf y$]{\includegraphics[height=1.1\sizeFig]{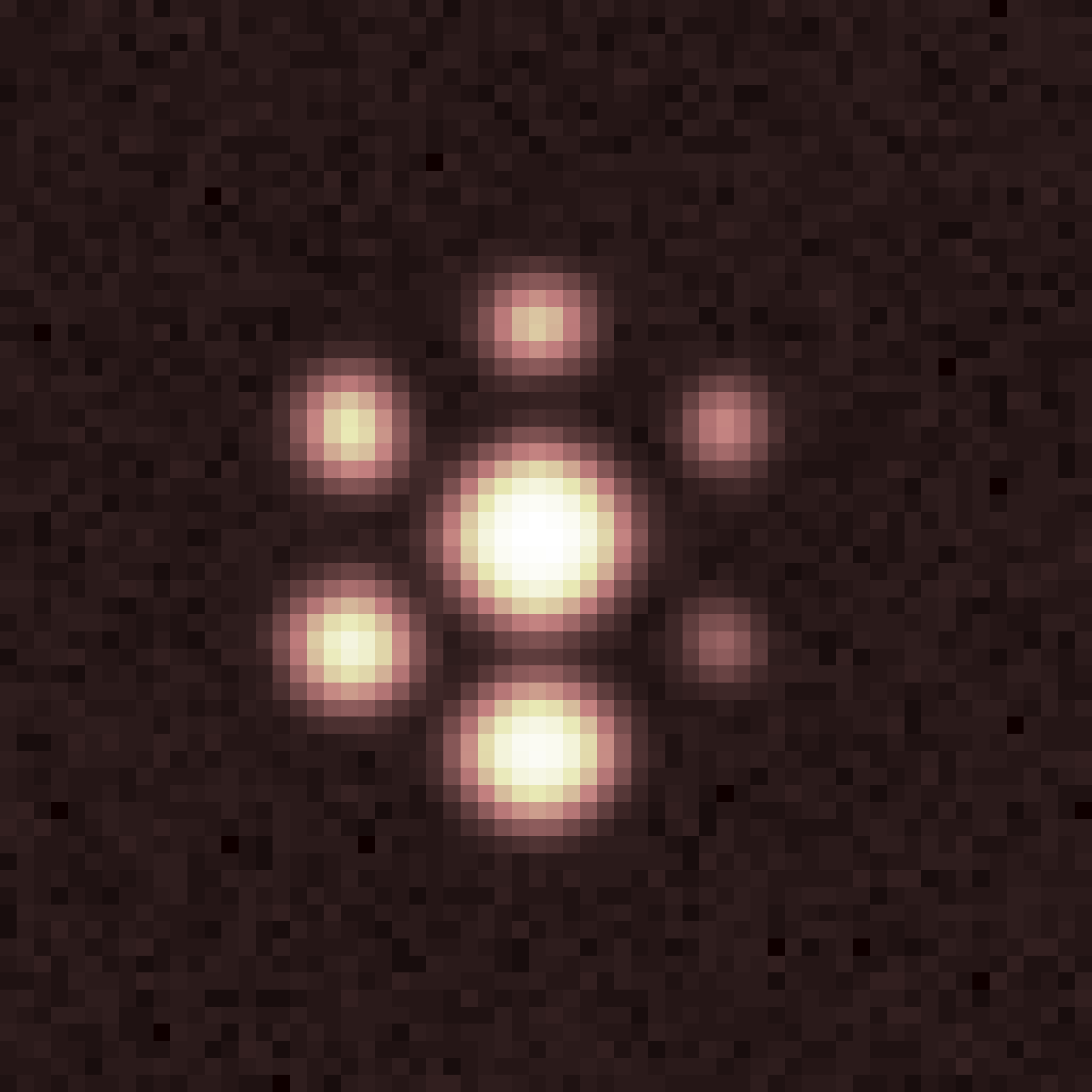}\begin{tikzpicture}[overlay]\draw (0,0) node {};\draw [red,very thick,dotted] (-0.5,1.15) -- (-1.85,1.15); \end{tikzpicture}} \hspace{\hspaceFig}
		\subfloat[\label{fig:results_synthetic_phantom_d}$\widetilde{\mbf h}_\text{BD}$]{\includegraphics[height=1.1\sizeFig]{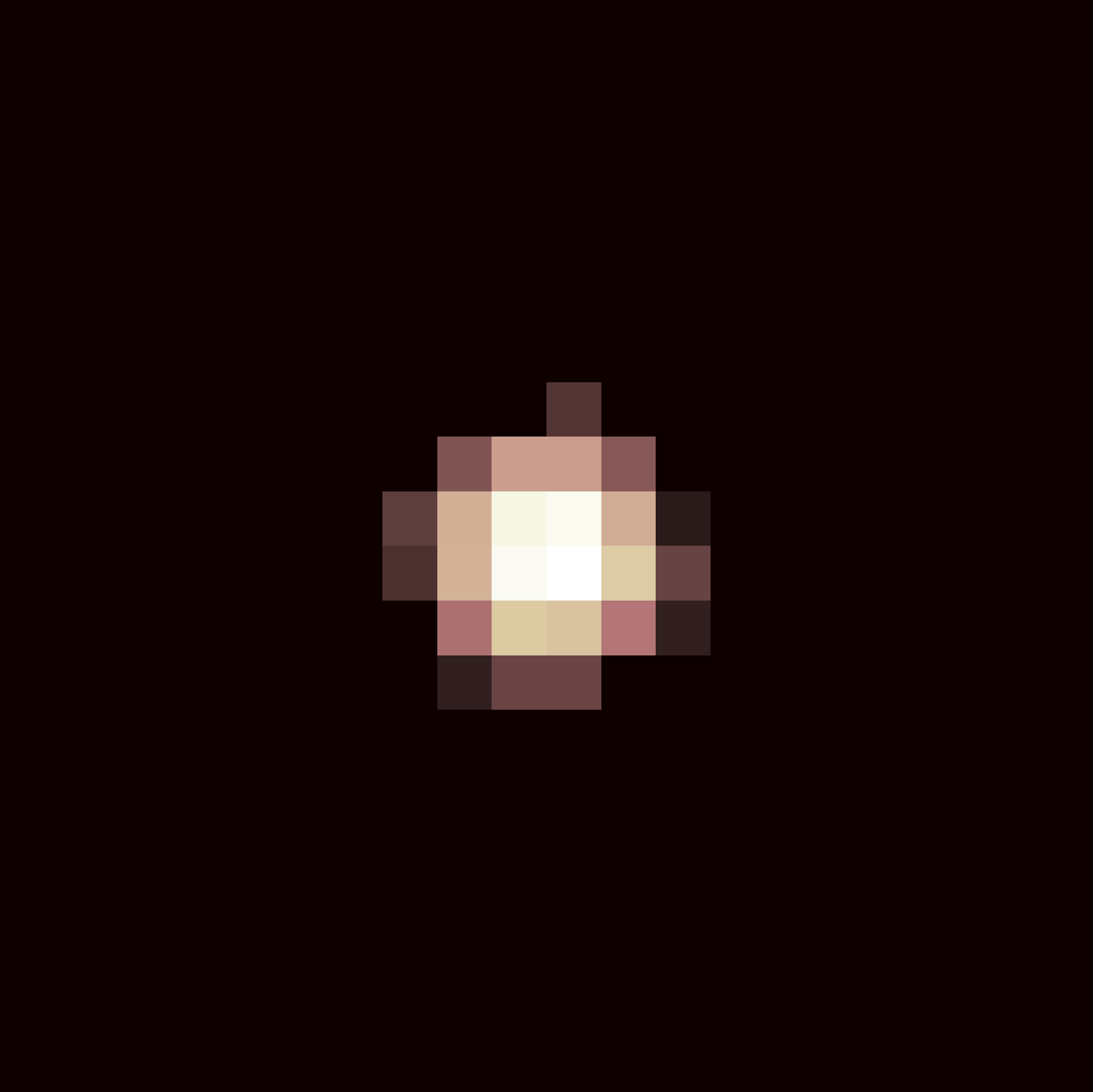}} \hspace{0.5mm}
		\subfloat[\label{fig:results_synthetic_phantom_e}$\widetilde{\mbf x}_{\text{NBD}}$ using $\widetilde{\mbf h}_\text{BD}$]{
		\includegraphics[height=1.1\sizeFig]{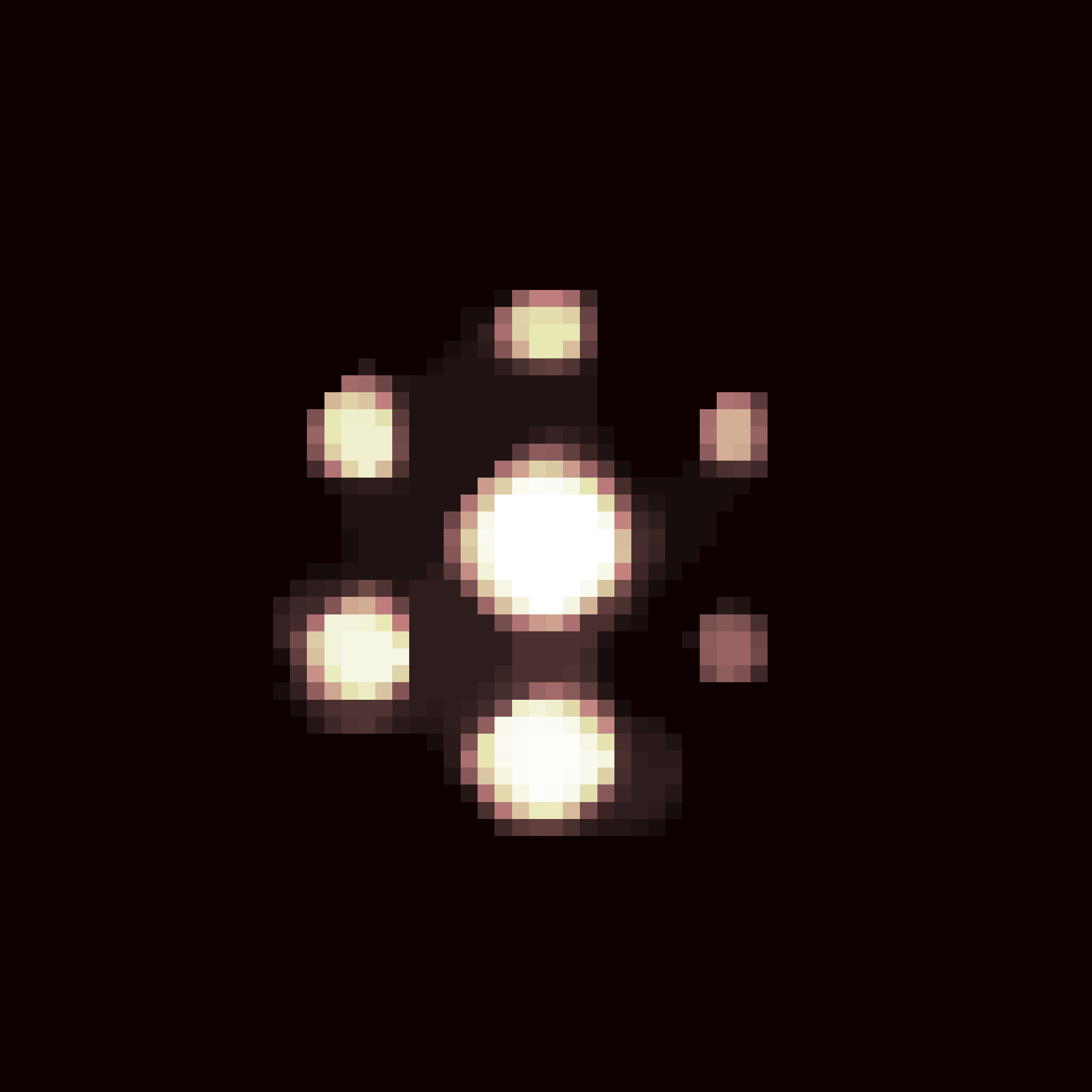}\begin{tikzpicture}[overlay]\draw (0,0) node {}; \draw [gray,very thick,solid] (-0.5,1.15) -- (-1.85,1.15);\end{tikzpicture}}\hspace{1mm}
		\subfloat[\label{fig:results_synthetic_phantom_f}1D profiles of $\mbf x$, $\mbf y$ and $\widetilde{\mbf x}_{\text{NBD}}$]{\begin{tikzpicture}\begin{axis}[	font = \footnotesize,height = 3.85cm,width = 0.3\textwidth,scale = 1,axis lines = middle,	xmin = 14, xmax = 50,ymin = 0, ymax = 1.1,every x tick/.style={color=black},every y tick/.style={color=black},legend cell align=left,xlabel style={anchor=center},ylabel style={align=center,yshift=2ex},ticks = none]	
				\pgfplotstableread{gt.dat}\gt
				\pgfplotstableread{xNB.dat}\xNB
				\pgfplotstableread{y.dat}\y
			\addplot+ [no marks, cyan, very thick, dashed] table[x expr=\coordindex, y expr = \thisrowno{1}, y index = 0]	{\gt};
			\addplot+ [no marks, gray, very thick, solid] table[x expr=\coordindex, y expr = \thisrowno{1}, y index = 0]	{\xNB};
			\addplot+ [no marks, red, very thick, dotted] table[x expr=\coordindex, y expr = \thisrowno{1}, y index = 0]	{\y}; \end{axis}\end{tikzpicture}}
	\caption{Results on synthetic phantom data (BSNR = 30\,dB). The original, observed and estimated images present a horizontal line at $n/2+1 = 33$. This corresponds to the profiles shown in Figure~\ref{fig:results_synthetic_phantom_f}, with $\mbf x$ in dashed cyan, $\mbf y$ in dotted red and $\widetilde{\mbf x}_\text{NBD}$ in solid gray.}
	\label{fig:results_synthetic_phantom}
\end{figure}
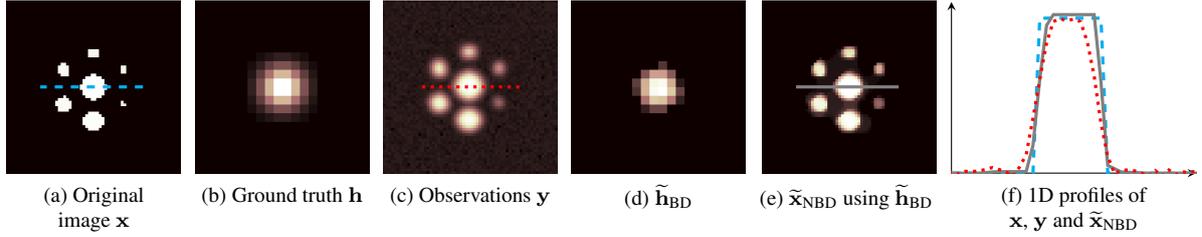

The synthetic image used for first validation of the method (see Fig.~\ref{fig:results_synthetic_phantom_a}) is similar to the image obtained from a phantom with cylindrical holes of known diameters filled with the same $^{18}$FDG activity concentration (1~mCi/100~mL). The discrete PSF is simulated by an isotropic Gaussian kernel with $\sigma = 1.3$ pixels (see Fig.~\ref{fig:results_synthetic_phantom_b}), \ie a PSF similar to those empirically estimated. The measurements are generated according to~\eqref{eq:discrete_model_vector} (see Fig.~\ref{fig:results_synthetic_phantom_c}). Noise variance $\sigma_\text{n}^2$ is chosen such that the blurred signal-to-noise ratio (SNR) defined as \mbox{BSNR = $10\log_{10} \text{var}(\mbf h \otimes \mbf x)/\sigma_\text{n}^2$}
is in $\{10, 20, 30, 40\}\,\text{dB}$. 
Quality of $\widetilde{\mbf x}$ and $\widetilde{\mbf h}$ is measured with the increase in SNR (ISNR) and the reconstruction SNR (RSNR), respectively~\cite{GDJ2016}. 

Real images were acquired on a Philips GEMINI-TF PET/CT scanner with an acquisition time of 15~min (see Fig.~\ref{fig:results_real_phantom_a}). The pixel size in transaxial slices is $2\times2$ mm$^2$. For comparison, we made a Gaussian approximation of the PSF ($\widetilde{\mbf h}_\text{Gaussian}$, see Fig.~\ref{fig:results_real_phantom_b}) by imaging a needle filled with $^{18}$FDG (3~mCi/mL). This PSF was used in the NBD scheme to deconvolve $\mbf y$, resulting in $\widetilde{\mbf x}_\text{Gaussian}$ (see Fig.~\ref{fig:results_real_phantom_c}).

Image size is $64 \times 64$ pixels ($N = 64^2$). For restoration, we consider that region $\Omega$ is composed of the five largest disks in the phantom mask, delineated thanks to the CT image (see Fig.~\ref{fig:results_real_phantom_d}).

\subsection{Results}
Table~\ref{tab:SyntheticResults_BSNR} presents the reconstruction results using the BD scheme for the considered levels of noise. They correspond to the average value over 10 trials. The results for BSNR = 30\,dB are depicted in Figs.~\ref{fig:results_synthetic_phantom_d}-\ref{fig:results_synthetic_phantom_f}. In terms of RSNR and visual observation, $\widetilde{\mbf h}_\text{BD}$ is close to the ground truth for different noise levels. When $\widetilde{\mbf h}_\text{BD}$ is used for the NBD,  $\widetilde{\mbf x}_\text{NBD}$ presents, as expected, a constant intensity in region $\Omega$. 
\begin{table}[ht]
\centering
	\begin{tabular}{ c | c | c }
		BSNR [dB] & RSNR($\widetilde{\mbf h}_\text{BD}$) [dB] & ISNR($\widetilde{\mbf x}_\text{BD}$) [dB] \\
		\hline
		40 & 19.47 & 11.38 \\
		30 & 16.34 & 10.18 \\
		20 & 12.60 & 7.82 \\
		10 & 6.20 & 4.71 \\
	\end{tabular}
\caption{BD results for synthetic phantom data and for different levels of noise.}
\label{tab:SyntheticResults_BSNR}
\end{table} 

\begin{figure}[b!]
	\centering
		\subfloat[\label{fig:results_real_phantom_a}Observations $\mbf y$]{\includegraphics[height=1.5\sizeFig]{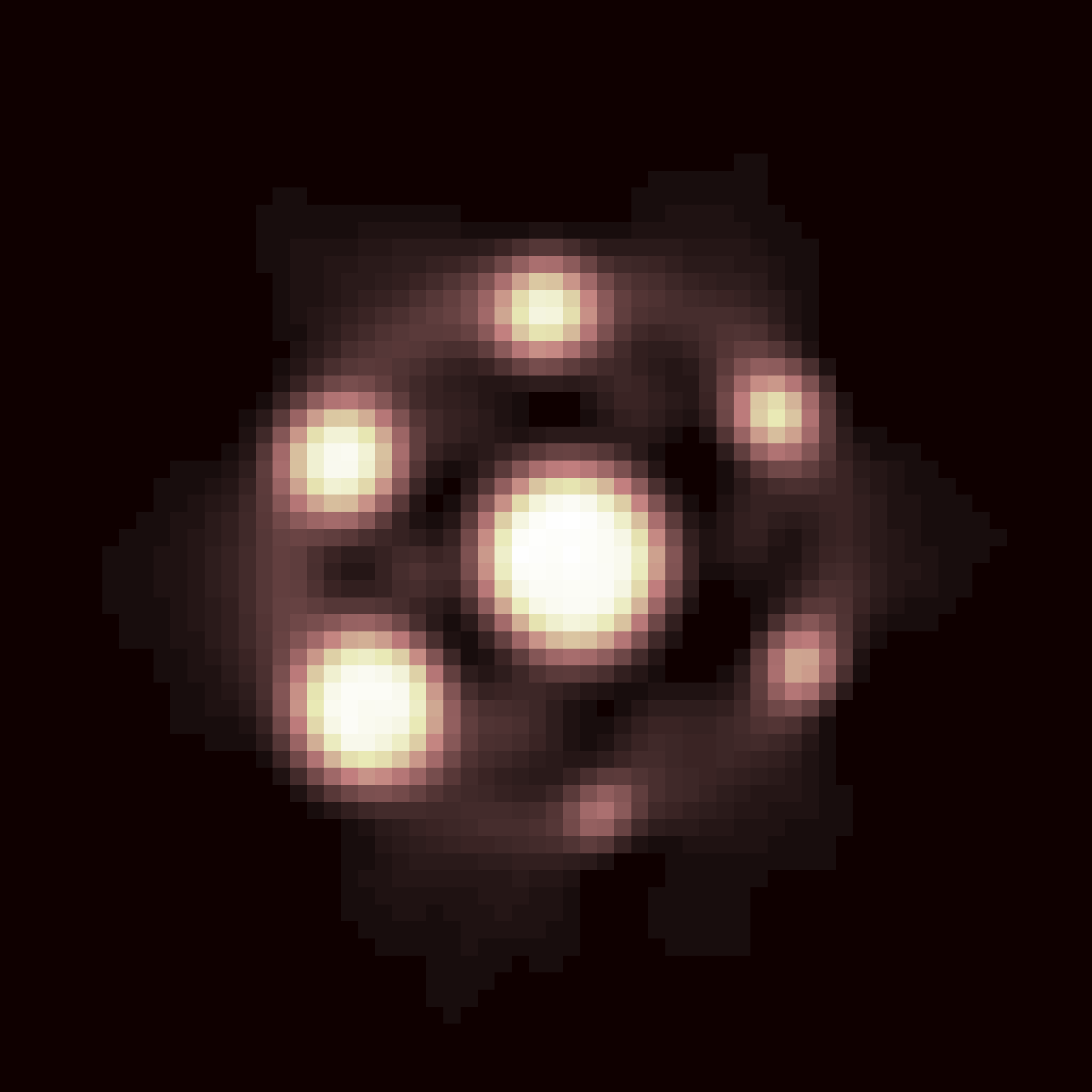}} \hspace{\hspaceFig}
		\subfloat[\label{fig:results_real_phantom_b}$\widetilde{\mbf h}_{\text{Gaussian}}$]{\includegraphics[height=1.5\sizeFig]{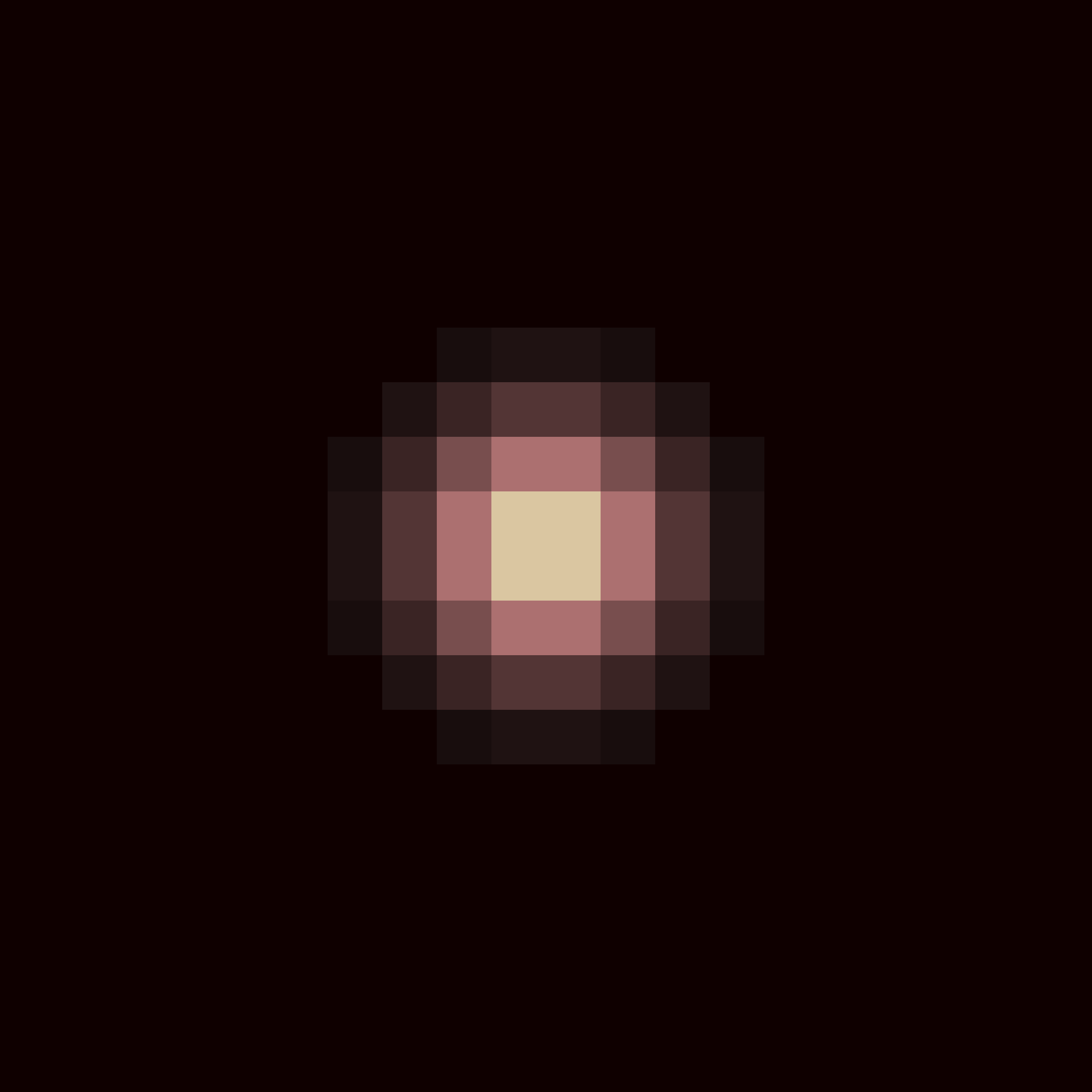}} \hspace{\hspaceFig}
		\subfloat[\label{fig:results_real_phantom_c}$\widetilde{\mbf x}_{\text{Gaussian}}$]{\includegraphics[height=1.5\sizeFig]{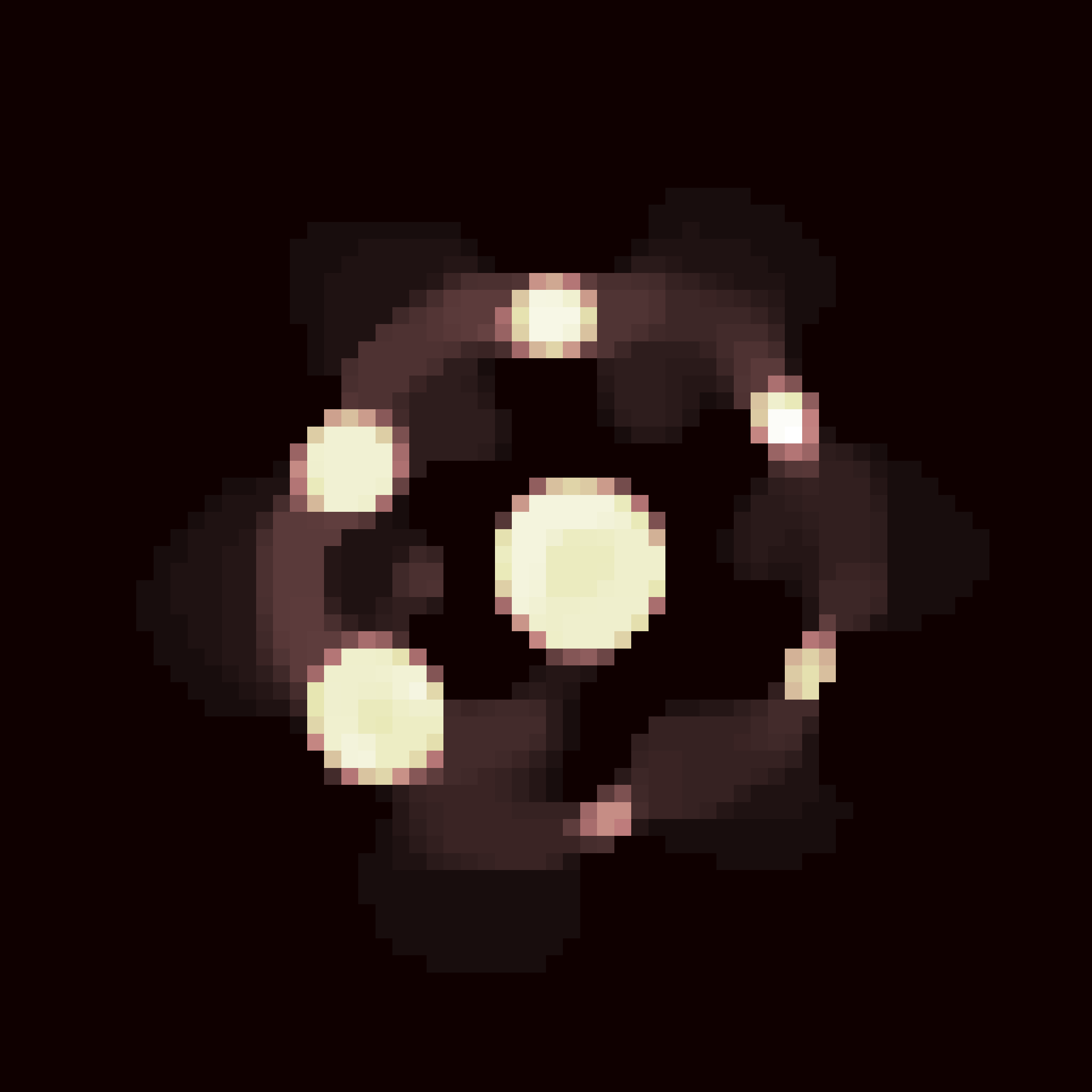}} \hspace{\hspaceFig}\\\vspace{-2mm}
		\subfloat[\label{fig:results_real_phantom_d}CT regions (red)]{\includegraphics[height=1.5\sizeFig]{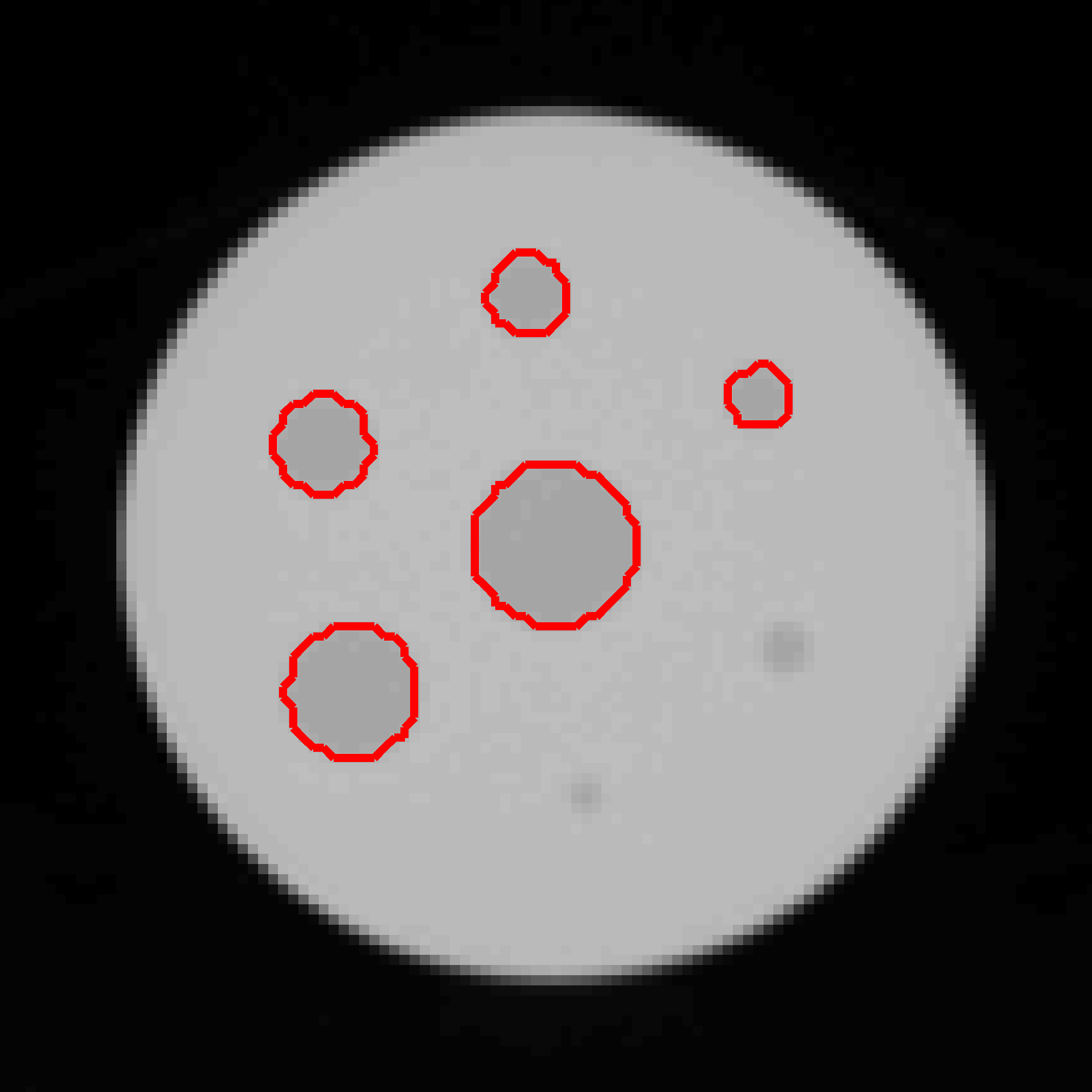}} \hspace{\hspaceFig}
		\subfloat[\label{fig:results_real_phantom_e}$\widetilde{\mbf h}_{\text{BD}}$]{\includegraphics[height=1.5\sizeFig]{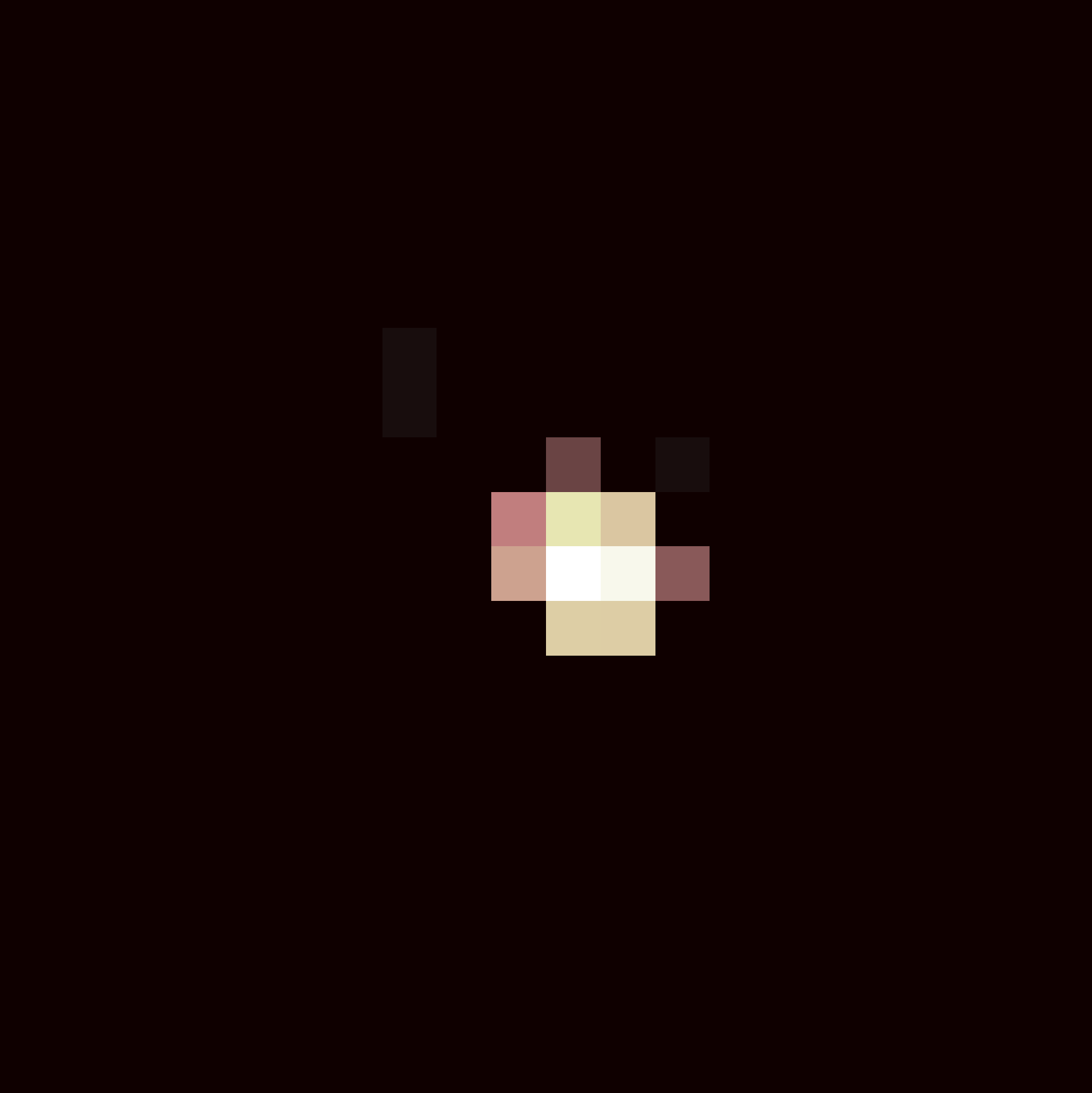}} \hspace{\hspaceFig}
		\subfloat[\label{fig:results_real_phantom_f}$\widetilde{\mbf x}_{\text{NBD}}$]{\includegraphics[height=1.5\sizeFig]{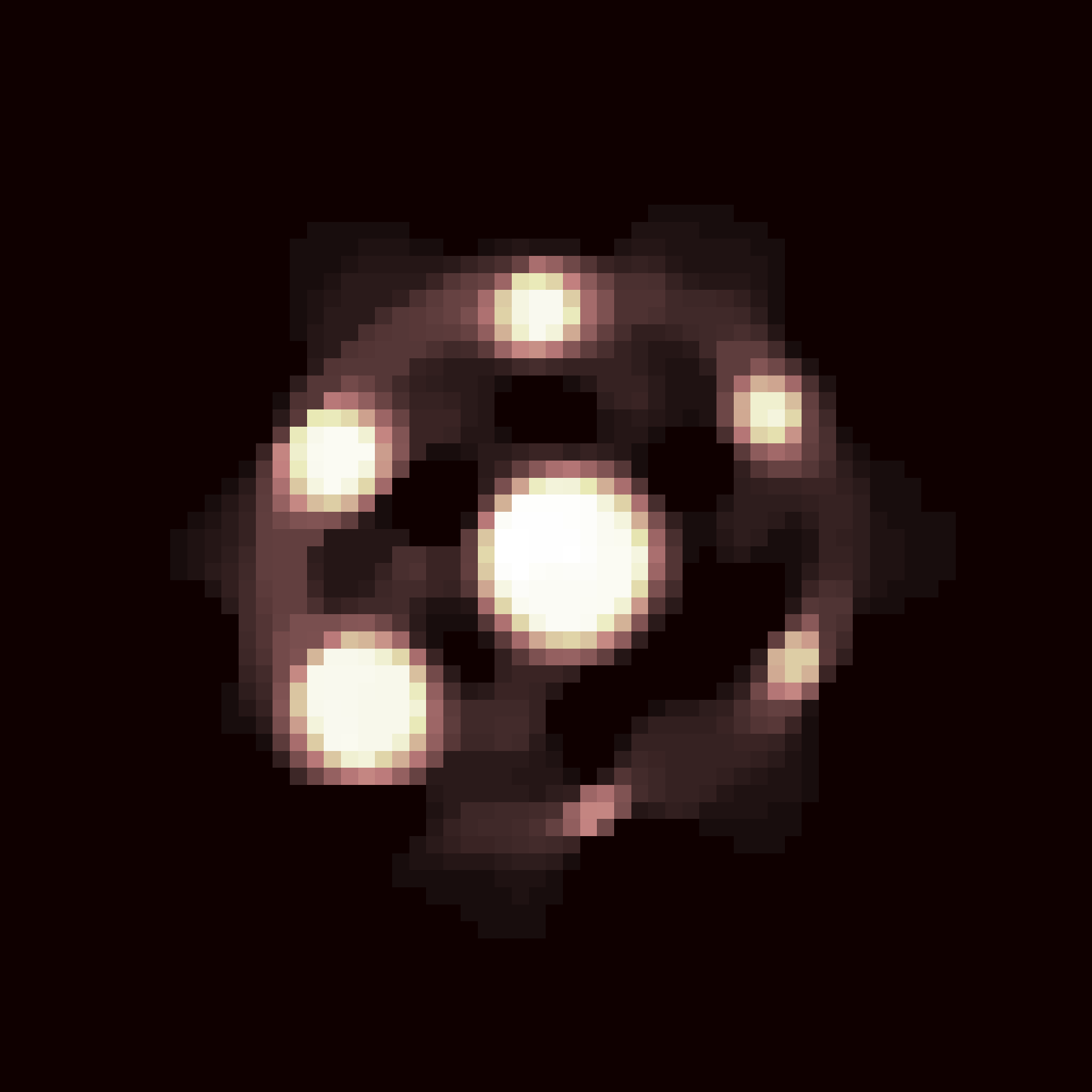}}
	\caption{Results on real phantom data.}
	\label{fig:results_real_phantom}
\end{figure}

Figure~\ref{fig:results_real_phantom} presents the results on real phantom data. The PSF $\widetilde{\mbf h}_\text{BD}$ is well-localized. Quality of the resulting NBD image is not as satisfactory as expected due to reconstruction artifacts.

\subsection{Perspectives}
In further research, we would like to improve: \emph{(i)} the forward model by considering the whole 3D image as well as Poisson noise statistics and \emph{(ii)} the assumptions and priors on the PSF by considering its spatial variation in the FOV and additional constraints (\eg sparsity).

\end{document}